\documentclass[letterpaper, 10 pt, journal, twoside]{packages/IEEEtran}

\usepackage[skip=2pt,font=footnotesize]{caption} 
\usepackage{multirow}
\usepackage{graphics} 
\usepackage{epsfig} 
\usepackage{times} 
\usepackage{amssymb}  
\usepackage{mathtools}
\usepackage{packages/commath}

\usepackage{amsthm}
\usepackage{graphicx}
\usepackage{subcaption}

\usepackage[colorlinks=true]{hyperref}
 \usepackage[font=small]{caption}

\usepackage{makecell}
\usepackage{tikz}

\usepackage{array}
\newcolumntype{M}[1]{>{\centering\arraybackslash}m{#1}}

\usepackage{colortbl}
\usepackage{tabulary}

\usepackage{amsmath,amsfonts} 

\usepackage{amsmath,amssymb,amsfonts}
\usepackage{graphicx}
\DeclareGraphicsExtensions{.eps,.pdf,.png,.jpg,.JPG,.gif}
\usepackage{bm}
\usepackage{url}
\usepackage{xcolor}

\newcolumntype{M}[1]{>{\centering\arraybackslash}m{#1}}

\usepackage{array}
\usepackage{booktabs}
\newcommand{\PreserveBackslash}[1]{\let\temp=\\#1\let\\=\temp}
\newcolumntype{C}[1]{>{\PreserveBackslash\centering}m{#1}}
\newcolumntype{R}[1]{>{\PreserveBackslash\raggedleft}p{#1}}
\newcolumntype{L}[1]{>{\PreserveBackslash\raggedright}p{#1}}

\usepackage[backend=bibtex,natbib=true,style=numeric-comp,sorting=none,firstinits=true,maxbibnames=99,url=false,doi=false]{biblatex}

\usepackage{xcolor}

\newcommand\rev[1]{{{#1}}}

\usepackage{packages/algorithm}
\usepackage{packages/algorithmic}\usepackage{array} 
\usepackage{adjustbox}
\usepackage{dashrule}

\newcommand{\bx}{\mathbf{x}}

\newcommand{\bK}{\mathbf{K}}
\newcommand{\bO}{\mathbf{O}}

\newcommand{\bT}{\mathbf{T}}

\newcommand{\bz}{\mathbf{z}}

\newcommand{\bC}{\mathbf{C}}

\newcommand{\bB}{\mathbf{B}}
\newcommand{\bH}{\mathbf{H}}

\newcommand{\bI}{\mathbf{I}}

\newcommand{\bD}{\mathbf{D}}

\newcommand{\bu}{\mathbf{u}}

\newcommand{\bS}{\mathbf{S}}
\newcommand{\bM}{\mathbf{M}}
\newcommand{\bF}{\mathbf{F}}

\newcommand{\br}{\mathbf{r}}

\usepackage{mathtools}

\allowdisplaybreaks 
\renewcommand{\baselinestretch}{0.96}
\begin{document}

\title{\Large \bf
Incremental Dense Reconstruction from Monocular Video \\ with Guided Sparse Feature Volume Fusion
}
\author{Xingxing Zuo$^{1,2}$, Nan Yang$^{1}$, Nathaniel Merrill$^{3}$, Binbin Xu$^{4,5,\dag}$, Stefan Leutenegger$^{1,2,4,6}$
%
	\thanks{$^1$ School of Computation, Information and Technology, Technical University of Munich, Germany. Email: \tt \small firstname.lastname@tum.de}%
    \thanks{$^2$ Munich Center for Machine Learning (MCML), Germany.}
    \thanks{$^3$ University of Delaware, USA.  Email: \tt \small 
     nmerrill@udel.edu}%
    \thanks{$^4$ Department of Computing, Imperial College London, United Kingdom.}%
    \thanks{$^5$ University of Toronto Robotics Institute, University of Toronto, Canada. Email: \tt \small binbin.xu@utoronto.ca}%
    \thanks{$^6$Munich Institute of Robotics and Machine Intelligence (MIRMI), Germany.}
    \thanks{$^\dag$ Corresponding author.}
}

\markboth{IEEE Robotics and Automation Letters. Preprint Version. April, 2023}{Zuo \MakeLowercase{\textit{et al.}}: Incremental Dense Reconstruction from Monocular Video}

\maketitle

\begin{abstract}
	\rev{Incrementally recovering 3D dense structures from monocular videos is of paramount importance since it enables various robotics and AR applications.}
	Feature volumes have recently been shown to enable efficient and accurate incremental dense reconstruction without the need to first estimate depth, but they are not able to achieve as high of a resolution as depth-based methods due to the large memory consumption of high-resolution feature volumes.
	This letter proposes a real-time feature volume-based dense reconstruction method that predicts TSDF (Truncated Signed Distance Function) values from a novel {\em sparsified} deep feature volume, which is able to achieve higher resolutions than previous feature volume-based methods, \rev{and is favorable in outdoor large-scale scenarios where the majority of voxels are empty}.
	An uncertainty-aware multi-view stereo (MVS) network is leveraged to infer initial voxel locations of the physical surface in a sparse feature volume.
	Then for refining the recovered 3D geometry, deep features are attentively aggregated from multi-view images at potential surface locations, and temporally fused. 
	%
	%
	%
	Besides achieving higher resolutions than before, 
	our method is shown to produce more complete reconstructions with finer detail in many cases.
	Extensive evaluations on both public \rev{and self-collected datasets} demonstrate a very competitive real-time reconstruction result for our method compared to state-of-the-art reconstruction methods \rev{in both indoor and outdoor settings}.
\end{abstract}


\begin{figure*} 
	\centering
	\includegraphics[width=0.9\textwidth]{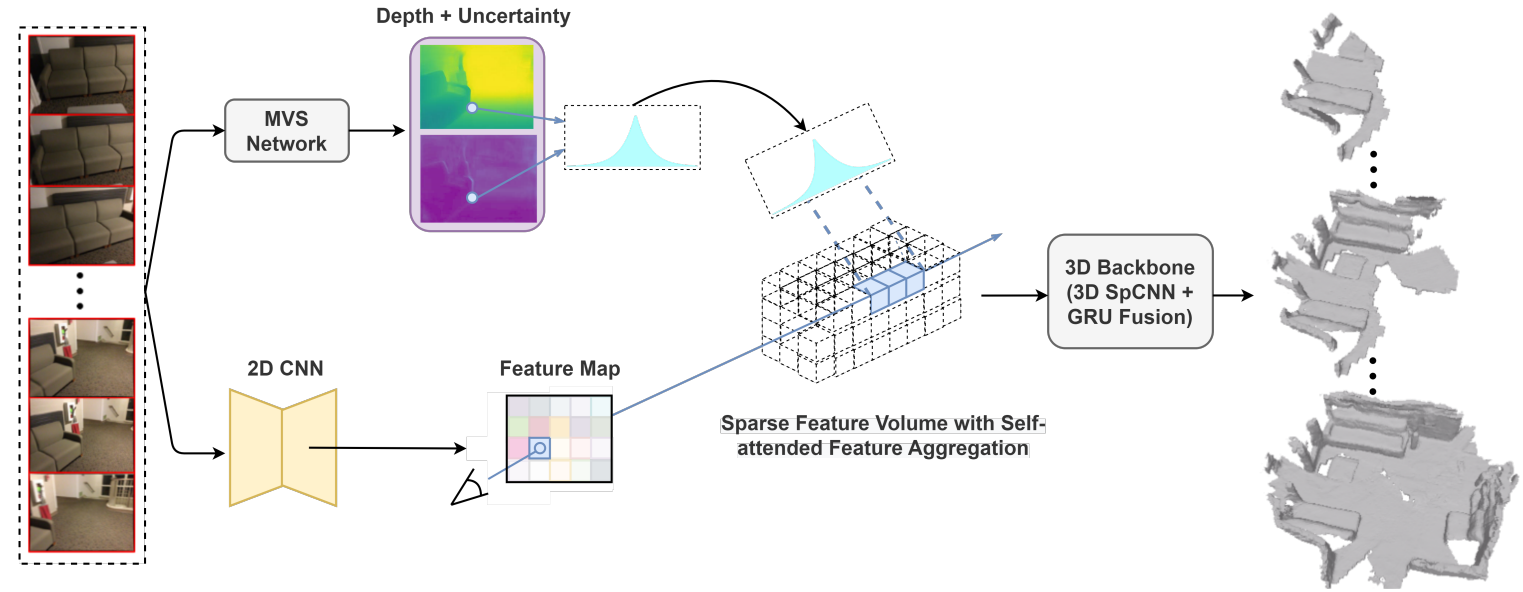}
	\caption{The overview of our proposed method. We leverage an MVS neural network for depth and depth uncertainty predictions, which provide an initial guess of the physical surface location. Then feature volume-based reconstruction pipeline allocate and aggregate deep features around the physical surface, formatting sparse feature volume. TSDF values can be directly regressed from the feature volume which is incrementally updated. Note that the pipeline involves three levels of feature volumes, which are omitted here for simplicity.
	}
	\label{fig:framework}
	\vspace*{-2em}
\end{figure*}

\section{Introduction}
\IEEEPARstart{D}{ense} reconstruction from video images with deep neural networks has attracted significant attention in recent years. Deep feature volume-based 3D scene reconstruction, regressing scene geometry directly from volumetric feature volume, has shown promising results~\cite{murez2020atlas, sun2021neuralrecon, bozic2021transformerfusion, stier2021vortx}, and has the potential to enable a wide range of robotic applications. The incremental variant~\cite{sun2021neuralrecon} can even achieve real-time performance on a desktop with commercial-level GPU.
Compared to predicting dense depth at multiple views and then fusing depths into a global 3D map, feature volume-based method back-project encoded 2D image features into 3D voxel grids, and directly regress truncated signed distance function (TSDF) from accumulated features across multiple image views, by using \rev{a} neural network composed of 3D convolutional layers and multiple layer perceptron (MLP) layers. Operation and prediction directly on the view-independent 3D feature volume have the advantage of capturing smoothness and 3D shape prior of the surface in the scene.
%

However, there are several drawbacks for existing feature volume-based methods~\cite{murez2020atlas, sun2021neuralrecon, bozic2021transformerfusion, stier2021vortx}.
Firstly, allocating features into all visible voxels along the whole rays cast from image pixels is cumbersome and redundant, which not only creates unnecessary confusion for network inference but also incurs excessive memory consumption and heavy computational burden. The ideal solution is only allocating image features into the relevant regions in 3D space, i.e., around the physical surface, for reconstruction. In the case that surface location is not certainly known, features can be allocated to the potential region where the surface is likely to locate.
Secondly, due to the memory and computation issue for the volumetric dense feature volume, existing methods~\cite{murez2020atlas, sun2021neuralrecon, bozic2021transformerfusion, stier2021vortx} are not capable of high-resolution reconstruction.
All of them have demonstrated to ability to perform 3D dense reconstruction with feature volume, using a voxel size of over 4cm at the finest level. It is acceptable but causes visible aliasing artifacts \rev{such as surfaces appearing overly smooth and lacking fine details.} Besides, the memory consumption of feature volume  grows cubically with the increased volumetric resolution, which hinders existing methods from scaling up to high-resolution and fine-detailed reconstruction. 

To address the aforementioned issues, we propose a novel guided sparse feature volume fusion method for real-time incremental scene reconstruction. 
Feature volumes are constructed fragment by fragment, and temporally fused into a global one. This incremental paradigm frequently updates the reconstructed 3D map, which favors real-time applications.
To maintain the sparsity of the feature volume for efficient reconstruction, we propose a method to selectively allocate features into only relevant voxels around the actual physical surface, which aims to avoid excessive memory and computation consumption. We firstly leverage an efficient MVS network to predict dense depth and depth uncertainty, which is used to select the sparse set of voxels to be aggregated for predicting the surface.
%
A self-attention mechanism~\cite{vaswani2017attention} is utilized for feature aggregations across multiple views,
and then 3D sparse convolutions are performed on the feature volume, followed by Gated Recurrent Unit (GRU)~\cite{chung2014empirical} to temporally fuse the feature volume fragment into the global one.
We also utilize traditional TSDF fusion~\cite{newcombe2011kinectfusion} with the available depths from MVS to generate a rough TSDF map, which is used as an additional feature channel to guide the feature volume-based reconstruction. The contribution of this paper can be summarized as follows:
\begin{itemize}
	\item We develop a real-time incremental reconstruction system for monocular video images based on novel sparse feature volume fusion.   
	\item We propose to utilize MVS neural networks to predict initial depth and depth uncertainty maps for efficient feature allocation into 3D voxels, which maintains the natural sparsity of the problem and allows our method to recover more fine-grained details and \rev{to work effectively in outdoor large-scale scenarios}. 
	\item The proposed method is verified on various datasets, and demonstrated to have
	competitive reconstruction accuracy compared to state-of-the-art methods, 
	and able to predict at a higher resolution than previous feature volume-based methods.
\end{itemize}

\section{Related Work}

Dense reconstruction from monocular videos has become increasingly accurate and robust with the emergence of deep learning. Our work is relevant to multi-view stereo (MVS) networks and feature volume-based scene reconstruction.

\subsection{Depth-based Reconstruction with MVS Networks.}
Given a set of monocular images with known poses and camera intrinsics, the goal of MVS networks is to infer the dense depth map of the reference frame using the provided multi-view information. 
Inspired by traditional plane-sweeping MVS methods~\cite{collins1996space},  a rich body of MVS neural networks first construct a plane-sweeping cost volume using the image intensity values~\cite{wang2018mvdepthnet, hou2019multi} or deep 2D features generated from images~\cite{yao2018mvsnet,im2019dpsnet,wang2021patchmatchnet,duzceker2021deepvideomvs}. Then the depth map of the reference image can be regressed from the cost volume through 2D convolutions~\cite{wang2018mvdepthnet, huang2018deepmvs, hou2019multi} or 3D convolutions~\cite{yao2018mvsnet,im2019dpsnet}.

Most of the methods that can be considered for real-time applications utilize 2D convolutions.
DeepMVS~\cite{huang2018deepmvs} exploits patch matching for plane-sweep volume generation, and incorporates both intra-volume and inter-volume feature aggregation across an arbitrary number of input images.
MVDepthNet~\cite{wang2018mvdepthnet} is an efficient network enabling real-time applications. It generates a cost volume directly from the warped image pixel values, and regresses the depth via a lightweight encoder-decoder network composed of 2D convolutions and skip connections.
GPMVS~\cite{hou2019multi} further extends MVDepthNet with the introduction of Gaussian Process (GP) prior to the bottleneck layer. 
The intuition is to leverage a pose kernel to measure the difference between camera poses, and to encourage similar poses to have similar latent variables in the bottleneck layer. With only a slight computation increment originating from the GP prior constraint, GPMVS can achieve much higher accuracy on dense depth regression over MVDepthNet.
SimpleRecon~\cite{sayed2022simplerecon} achieves high accuracy for depth prediction by including extra information in the cost volume via parallel MLP reduction of readily-available metadata~-- such as dot product of features across multiple views, back-projected rays from pixels, pose distances, and a validity mask~-- but incurs a higher computational cost than competing methods.
In order to lift the 2D depth images into a 3D volumetric representation, traditional TSDF fusion~\cite{curless1996volumetric, newcombe2011kinectfusion} is typically utilized by these methods.
%

%
MVSNet~\cite{yao2018mvsnet} 
constructs a 3D cost volume based on the variance of deep features, and further regularizes and regresses the depth map of the reference image via 3D convolutions -- which enables higher accuracy than 2D convolutions but pays the cost of higher computation and memory consumption.
To alleviate this issue, and allow depth prediction at higher resolutions, Gu et al. propose a cascade 3D cost volume~\cite{gu2020cascade} -- narrowing the depth range of cost volume gradually from coarse to fine scales.
PatchmatchNet~\cite{wang2021patchmatchnet} imitates the traditional PatchMatch method~\cite{barnes2009patchmatch} by an end-to-end trainable architecture, which reduces the number of 3D convolutions needed in cost volume regularization and allows prediction at an even higher resolution than the cascade method~\cite{gu2020cascade}.
While the accuracy and efficiency of the MVS methods utilizing 3D convolutions is ever increasing,
they are still mostly amenable to offline processing.
\subsection{Feature Volume-Based Scene Reconstruction.}
SurfaceNet~\cite{ji2017surfacenet} is among the first works to directly predict surface probability from 3D voxelized colored volumes from two image views using a 3D convolutional network. For generating the 3D voxelized colored volume, all the pixels on images are projected into 3D through the known camera intrinsics and extrinsics. Two colored volumes are then concatenated along the color channel, and regularized by 3D convolutions.
Atlas~\cite{murez2020atlas} extends this idea by replacing the colored volume with more informative deep feature volumes, and further enables an arbitrary number of multi-view images. Constructed 3D volumetric deep feature volumes across multiple views go through average pooling before being fed into the 3D convolutions.
TransformerFusion~\cite{murez2020atlas} and Vortx~\cite{stier2021vortx} exploit transformer-based attention mechanism for aggregating features from multiple image views instead of average pooling. TransformerFusion~\cite{murez2020atlas} also leverages the predicted attention weights to select the most relevant information for fusion.
In order to alleviate the effects of occlusion in the aggregation of multi-view image features, Vortx~\cite{stier2021vortx} predicts the projective occupancy probabilities, which are used as weights to produce the aggregated feature in the volume.

Note that none of the aforementioned feature volume-based methods are aiming for real-time applications. 
TransformerFusion~\cite{murez2020atlas} processes every image frame one by one, and gradually selects a certain number of the most relevant feature measurements for every voxel grid based on attention weights. Besides, expensive dense 3D convolutions are utilized to deal with the feature volume. Those mentioned assignment choices make TransformerFusion far from real-time \rev{capable}.
Vortx~\cite{stier2021vortx} does not work in an incremental way, and it predicts the TSDF map from the final integrated feature volume with the aggregated features from certain selected image views in the whole video stream. 

In contrast to the above methods, we focus on real-time incremental reconstruction based on feature \rev{volumes}.
The closest work to our method is NeuralRecon~\cite{sun2021neuralrecon}, which is also a baseline of our method. It performs feature volume-based reconstruction fragment by fragment at the first phase, which appears similar in spirit to active sliding windows in traditional SLAM methods~\cite{mourikis2007multi, 
	leutenegger2013keyframe, engel2017direct, zuo2019lic, 
}. In order to get a globally consistent reconstruction, NeuralRecon further adopts GRU-Fusion at the second phase to fuse the fragment feature volumes over time, which can be regarded as an alternative to the conventional TSDF fusion~\cite{curless1996volumetric, newcombe2011kinectfusion}. 
NeuralRecon does not have access to pick out the most relevant features from the whole video before the feature volume fusion and processing, thus it is supposed to have inferior performance than the full-batch methods~\cite{murez2020atlas,stier2021vortx}. 
Distinct from all the existing feature volume-based  methods that unproject and allocate features into the voxel grids along the whole rays in feature volume, we leverage MVS for rough dense depth predictions -- which allows us to allocate features to sparse voxels around the physical surfaces only. The retained sparsity keeps the memory consumption low and further enables high-resolution feature volume for scene reconstruction.

\section{Methodology}\label{sec:methodology}

We take as input a fragment sequence of $N$ keyframe images $\{\bI_k\}_{k=0}^{N-1}$ along with their corresponding poses $\{ \bT_k\}_{k=0}^{N-1}$ and camera intrinsics $\{\bK_k\}_{k=0}^{N-1}$, $N=9$ is used in our work. Following~\cite{duzceker2021deepvideomvs, sun2021neuralrecon,}, we utilize a 2D feature extraction network composed of an MnasNet encoder~\cite{tan2019mnasnet} and feature pyramid network (FPN)~\cite{lin2017feature} style decoder. 
We unproject extracted features into a 3D aggregated feature volume representation and directly regress the sparse TSDF values from \rev{the} feature volume. The key insight in our method
is the use of {\em depth priors} to construct a feature volume that is sparse from the very start~--
allowing our 3D network to focus on the surface from the very beginning without wasting effort in allocating and processing dense volumes.
An overview of our system can be seen in Fig.~\ref{fig:framework}.

\subsection{MVS-Guided Sparse Feature Allocation}
Unlike existing feature volume-based methods~\cite{murez2020atlas, sun2021neuralrecon, stier2021vortx} that allocate dense feature volumes from unprojected features,
we utilize depth priors (depth map and its uncertainty) to allocate feature volume locations only where the physical surface is likely located. 
To get the depth priors of every keyframe in the fragment for efficient feature allocation, 
we leverage GPMVS~\cite{hou2019multi} due to its appealing efficiency and adequate accuracy. Due to the modularity of our design, other MVS methods like~\cite{duzceker2021deepvideomvs, sayed2022simplerecon} could also be applied to our method.

\subsubsection{MVS-Based Depth and Uncertainty Prediction}\label{subsec:mvs}

%
GPMVS~\cite{hou2019multi} predicts the inverse depth map $\hat{\bD}_{L_i}^{-1}$ at four scales $i \in \{0, 1, 2, 3\}$, and applies supervision at the four levels by resizing the ground truth inverse depth $\bD_{L_i}^{-1}$. Note that we use $\;\hat{\cdot}\;$ to denote the estimated/predicted variables and, for ease of notation, assume operations (inverse, $\exp$, etc.) to be element-wise throughout this paper. 
We augment the GPMVS network architecture to enable the prediction of dense depth uncertainty $\hat{\bB}$ which is parameterized by $\hat{\bB} = \exp(\hat{\bB}_{\mathrm{log}})$ to ensure a positive uncertainty value. To predict $\hat{\bB}_{\mathrm{log}}$ we simply duplicate the last three layers of the decoder in GPMVS to create a shallow second decoder head at the highest resolution.
%
%
For the highest GPMVS resolution $L_0$, following~\cite{bloesch2018codeslam, zuo2021codevio}, we apply the Laplacian maximum likelihood estimator (MLE) loss to enforce that the predicted uncertainty tightly bounds the true prediction error:
%
%
\begin{align}
	\label{eq:loss_mv}
	\mathcal{L}^{(0)}_{\mathrm{mvs}} =
	\frac{1}{|\Omega|}
	\sum_{\bu \in \Omega}
	\frac{|\hat{D}_{L_0}^{-1}(\bu) - D_{L_0}^{-1}(\bu)|}{\hat{B}(\bu)} 
	+ \log{\hat{B}(\bu)}
\end{align}
where $\Omega$ is the set of pixels with valid ground truth.
For the other three resolutions with no uncertainty prediction, we use the standard $\ell_1$ loss. The losses $\mathcal{L}^{(i)}_{\mathrm{mvs}}$ from all resolutions are added together with equal weights, and mean reduction is used across batches.
%
%
For the remainder of this work, we will only use the (inverse) depth at the highest resolution, ${L_0}$, and omit the respective subscript for brevity.

We leverage linear uncertainty propagation to convert the uncertainty of inverse depth to the uncertainty of depth:
\begin{align}
	\hat{\bC} = \hat{\bD}^{2} \odot \hat{\bB}
\end{align}
\rev{where $\odot$ is the element-wise product.}

\subsubsection{Sparse Feature Allocation}\label{subsec:feat_allocation}

\begin{figure} [!bth]
	\centering
	\includegraphics[width=0.8\columnwidth]{%
		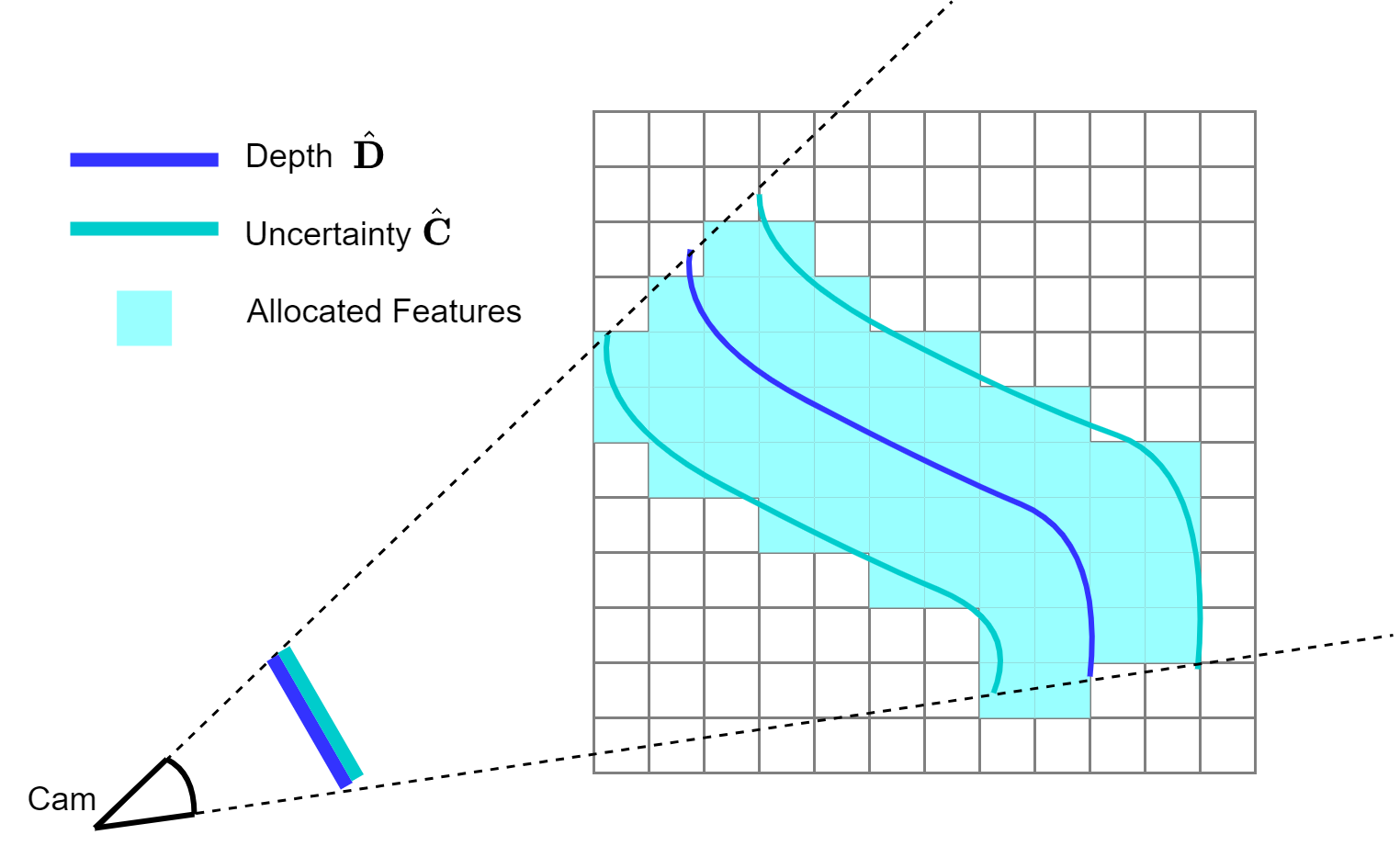}
	\caption{
		2D illustration of sparse feature allocation with the predicted depth map $\hat{\bD}$ and accompanying uncertainty map $\hat{\bC}$.
	}
	\label{fig:sparseallocation}
	\vspace*{-.4em}
\end{figure}

Here we present the scheme to do sparse feature allocation based on casting the predicted depth and uncertainty (Sec.~\ref{subsec:mvs}) into sparse voxels.
Fig. \ref{fig:sparseallocation} shows a visual representation of how the sparse feature volume is constructed.
For each keyframe in a fragment, we utilize the predicted depth $\hat{\bD}$ and uncertainty map $\hat{\bC}$ to determine where we should allocate voxels in the sparse feature volume. 
Specifically, we only allocate voxels in the feature volume that lie along the ray of a pixel with positive predicted depth value and within the range of uncertainty bounds $[\hat{\bD}- s \hat{\bC}, \hat{\bD}+ s \hat{\bC}]$.
In our experiments, we use $s=2$.

%

\subsection{Multi-view Feature Aggregation via Self Attention}
Deep image features at three resolutions are obtained by the feature extraction backbone, consisting of an efficient variant of \rev{MnasNet} encoder~\cite{tan2019mnasnet} followed by a feature pyramid network (FPN)~\cite{lin2017feature}. 
A 3D voxel location can be observed by multiple images from different viewpoints. We perform the feature aggregations in a content-aware way via multi-head self-attention~\cite{vaswani2017attention} at three scales. We first project a 3D voxel location into image planes across different views by the known camera poses and camera intrinsics, and fetch features for this specific voxel from the extracted multiple-scale feature maps through differentiable bilinear interpolation. Since the feature aggregations procedure at different scales is the same, we exemplify it at one scale. The fetched features from $N$ views, $\bF_{\mathrm{bp}} \in \mathbb{R}^{N \times C}$, with zero padding for the invisible views, and visibility binary mask $\bM \in \mathbb{R}^{N}$ are the input to the self-attention based feature aggregation module, which outputs the content-aware features with viewpoint and data dependencies:
\begin{align}
	\bF_{\mathrm{attn}} = f_{\mathrm{attn}}(\bF_{\mathrm{bp}}, \bM_{\mathrm{attn}})
\end{align}
where the output feature $\bF_{\mathrm{attn}} \in \mathbb{R}^{N \times C}$ has the same feature channel with the input.
Following~\cite{stier2021vortx}, we realize the above self-attention aggregation module $f_{\mathrm{attn}}(\cdot)$ by two transformer layers following the original transformer pipeline~\cite{vaswani2017attention}. Each layer includes a multi-head attention mechanism (\rev{two heads in our implementation}), as well as layer normalization, linear layers with ReLU activation, and residual connections. All the query, key, and value features originate from the same input features in the multi-head attention mechanism. In practice, we use two heads for the multi-head attention module. The aggregated features $\bF_{\mathrm{attn}}$ are simply averaged to generate a single feature vector $\bF$  for the specific voxel.

\subsection{Fragment Reconstruction from Sparse Feature Volume}
\rev{After aggregating features from different views, we obtain a single feature vector within each non-empty voxel. We then directly regress the TSDF value of the voxel from this feature vector. }
With the predicted dense depth maps of keyframes in the fragment, it is handy to perform conventional TSDF-fusion~\cite{newcombe2011kinectfusion} and get the TSDF values and weights for every voxel inside the chunk. The TSDF value and weight are concatenated with the averaged image feature for subsequent 3D sparse convolutions~\cite{tang2020searching}. The final TSDF values of the chunk can be directly predicted from the feature volume by an MLP layer.

\subsection{Fragment to Global Fusion}
We follow NeuralRecon~\cite{sun2021neuralrecon} to fuse the fragment feature volume  into a global feature volume incrementally via GRU fusion~\cite{chung2014empirical}.
For a feature vector $\bF_t$ originating from current fragment in the feature volume at time instant $t$, we fuse it with the historical feature $\bH_{t-1}$ at the same voxel location by GRU fusion. 
We observe that the volume resulting from traditional TSDF fusion with the predicted MVS depth is an additional useful feature for the network predicting the TSDF volume.
Thus, the fused TSDF values and weights, $\mathbf{S}_t$ and $\mathbf{S}_{Wt}$, from fusing the MVS depths are concatenated with features in order to guide the fusion process. After the concatenation with increased feature dimensions, we leverage single-layer MLPs for feature dimension reduction.
\begin{subequations}
	\begin{align}
		\mathbf{H}'_{t-1} &= \operatorname{MLP}_H\left(\left[\mathbf{H}_{t-1}, \mathbf{S}_t, \mathbf{S}_{Wt}\right]\right)\\
		\mathbf{F}'_{t} &= \operatorname{MLP}_F\left(\left[\mathbf{F}_{t}, \mathbf{S}_t, \mathbf{S}_{Wt}\right]\right)\\
		\bz_t &=\operatorname{sigmoid}\left(\operatorname{SpConv}\left(\left[\mathbf{H}'_{t-1}, \mathbf{F}'_t\right]\right)\right) \\
		\br_t &=\operatorname{sigmoid}\left(\operatorname{SpConv}\left(\left[\mathbf{H}'_{t-1}, \mathbf{F}'_t\right]\right)\right) \\
		\breve{\mathbf{H}}_t &=\tanh \left(\operatorname{SpConv}\left(\left[\br_t \odot \mathbf{H}'_{t-1}, \mathbf{F}'_t\right]\right)\right) \\
		\mathbf{H}_t &=\left(\mathbf{I}-\bz_t\right) \odot \mathbf{H}'_{t-1}+\bz_t \odot \breve{\mathbf{H}}_t
	\end{align}
\end{subequations}
where $\bz_t$ is the update gate vector, $\br_t$ the reset gate vector, $\left[ \cdot, \cdot \right]$ the concatenation operator. $\operatorname{SpConv}$ denotes the sparse point-voxel convolution operation~\cite{tang2020searching}. With the above GRU fusion, we can temporally fuse features and keep updating the visible feature volumes at three scales.

\subsection{Implementation Details}
We maintain three levels of feature volumes and regress the TSDF values $\bS_{L_i}$ from them in a coarse to fine manner. In order to further sparsify the feature volume for the proceeding scales, we also predict occupancy values $\bO_{L_x}$ from feature volumes at all scales with simple MLP layers. If the occupancy prediction of a voxel at a coarser scale is lower than a threshold (0.5), that voxel is redeemed as empty and will not be involved in feature allocation and prediction at finer scales~\cite{sun2021neuralrecon}. 
Overall, the training loss for regressing TSDF and occupancy at a single resolution $i$ is:
\begin{align}
	\label{eq:loss_recon}
	\mathcal{L}_{\mathrm{recon}}^{(i)} =
	\frac{1}{|\Lambda|}
	\sum_{\bx \in \Lambda}
	\bigg(  & \lambda_1  {|{\rm{logt}}(\hat{S}_{L_i}(\bx)) 
		- {\rm{logt}}({S}_{L_i}(\bx))|}  \bigg. \notag \\
	\bigg. & + \lambda_2 {\rm{BCE}}({\hat{O}_{L_i}(\bx), O_{L_i}(\bx)})  \bigg)
\end{align}
where $\Lambda$ is the set of voxels with valid ground truth, and ${\rm{logt}}(\bS_{L_i}) = {\rm{sign}}(\bS_{L_i})\log(|\bS_{L_i} + 1|)$ denotes the log-transform~\cite{sun2021neuralrecon,dai2020sg}, and $\rm{BCE}$ denotes the binary cross-entropy (BCE) loss.
\rev{We have $\lambda_1 = \lambda_2$ for balancing the two loss terms in our training.}
We add the losses $\mathcal{L}_{\mathrm{recon}}^{(i)} $ at each scale and apply mean reduction over batches.
%

%

The input images are at resolution $640\times480$, while the features at three levels are fetched from feature maps at resolutions $320\times240$, $160\times120$, $80\times60$ with channels $24, 40, 80$, respectively.
GPMVS requires input images with resolution $320 \times 256$, which are obtained by bilinear interpolation-based downsampling. Dense depth maps are downsampled via nearest neighbor to better preserve sharp edges. We utilize the TorchSparse~\cite{tang2020searching} implementation of sparse 3D convolutions in our method.

\section{Experiments}\label{sec:exp}

\begin{table}[!t]
	\centering 
	\caption{3D geometry metrics evaluated on ScanNet test split. 
		Note that the methods under the middle line are feature-volume-based incremental reconstruction methods, while the others are not.}
	\label{tab:scannet-3d}
	\resizebox{\columnwidth}{!}{
		\begin{tabular}{cccccc}
			\Xhline{3\arrayrulewidth}
			Method       & Comp $\downarrow$ & Acc $\downarrow$   & Recall $\uparrow$ & Prec $\uparrow$   & F-score  $\uparrow$   \\ \hline
			COLMAP \cite{schonberger2016pixelwise} & 0.069   & 0.135   & 0.634   & 0.505   & 0.558 \\ 
			MVDepthNet \cite{wang2018mvdepthnet}     &  \textbf{0.040}   & 0.240   & 0.831   & 0.208   & 0.329    \\ 
			DPSNet \cite{im2019dpsnet} & {0.045}   & 0.284   & \textbf{0.793}   & 0.223   & 0.344  \\
			GPMVS \cite{hou2019multi} & 0.105  & 0.191   & 0.423  & 0.339   & 0.373     \\ 
			Atlas \cite{murez2020atlas} & 0.083  & 0.101   & 0.566 & 0.600   & 0.579  \\ 
			Vortx \cite{stier2021vortx} & 0.081 & \textbf{0.062} & 0.605 & \textbf{0.689} & \textbf{0.643} \\
			\hline
			\rowcolor[gray]{0.902}
			NeuralRecon \cite{sun2021neuralrecon} & 0.137   & \textbf{0.056}  & 0.470   & \textbf{0.678}  & 0.553 \\
			\rowcolor[gray]{0.902}
			Ours & \textbf{0.110} & 0.058 & 0.505 & 0.665 & 0.572 \\
			\rowcolor[gray]{0.902}
			\rev{Ours (High Reso)} & 0.116 & \textbf{0.056} & \textbf{ 0.525} & 0.675 & \textbf{0.589} \\
			\Xhline{3\arrayrulewidth}
		\end{tabular}
	}
\end{table}

\begin{table}[!t]
	\centering
	\caption{2D depth metrics evaluated on ScanNet test split. Note that the methods under the middle line are feature-volume-based incremental reconstruction methods, while the others are not.}	
	\label{tab:scannet-2d}
	\resizebox{\columnwidth}{!}{
		\begin{tabular}{ccccccccc}
			\Xhline{3\arrayrulewidth}
			Method                                        & Abs-rel $\downarrow$ & Abs-diff $\downarrow$ & Sq-rel $\downarrow$ & RMSE $\downarrow$ & $\delta < 1.25$ $\uparrow$& Comp $\uparrow$ \\ \hline
			COLMAP \cite{schonberger2016pixelwise}                                        & 0.137          & 0.264          & 0.138          & 0.502          & 83.4                    & 0.871           \\ 
			MVDepthNet \cite{wang2018mvdepthnet}                                   & 0.098          & 0.191          & 0.061          & 0.293          & 89.6                    & 0.928           \\ 
			DPSNet \cite{im2019dpsnet}                                       & 0.087          & 0.158          & \textbf{0.035}          & 0.232          & 92.5           & 0.928           \\ 
			GPMVS \cite{hou2019multi}  &   0.088 &  0.206 & 0.053 & 0.359 & 90.3 & 0.928     \\ 
			Atlas \cite{murez2020atlas}    & 0.065          &
			0.123          & 0.045          & 0.249          & 92.4                    & \textbf{0.986}  \\
			Vortx \cite{stier2021vortx} & \textbf{0.057} & \textbf{0.090} & \textbf{0.035} & \textbf{0.197} & \textbf{93.9} & 0.951 \\ 
			\hline
			\rowcolor[gray]{0.902}
			NeuralRecon \cite{sun2021neuralrecon}                                       & 0.064 & 0.097 & 0.036 & 0.191 & 93.5           & 0.888           \\ 
			\rowcolor[gray]{0.902}
			ours & 0.057 & 0.092 & {0.030} & 0.183 & 94.2 & \textbf{0.913} \\
			\rowcolor[gray]{0.902}
			\rev{ours (High Reso)} & \textbf{0.052} & \textbf{0.087} & \textbf{0.025} & \textbf{0.175} & \textbf{94.8}& 0.906 \\
			\Xhline{3\arrayrulewidth}
		\end{tabular}
	}
\end{table}

\begin{figure*}
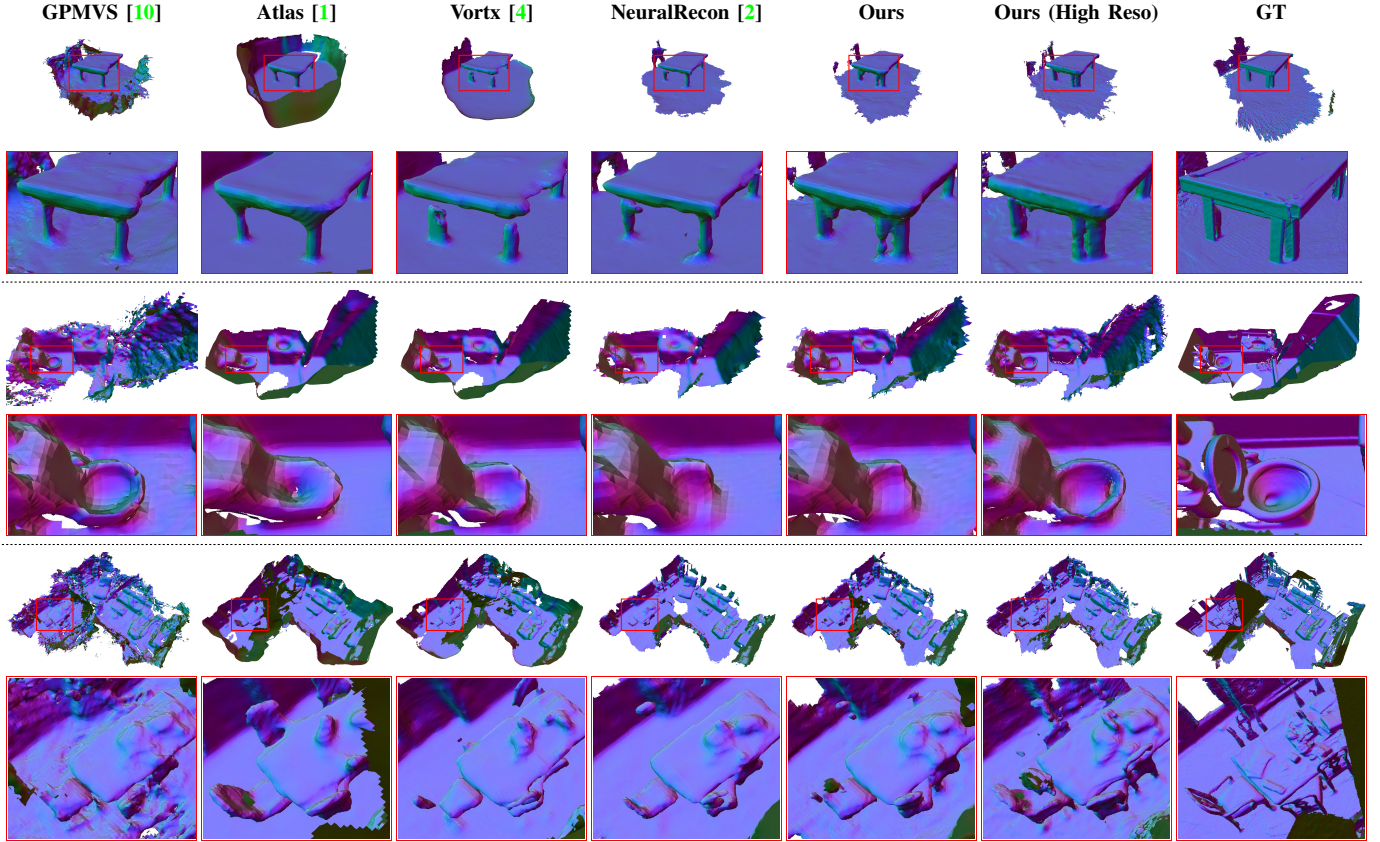

\newcommand{\w}{.14\textwidth}
\newcommand{\imgw}{2088.0}
\newcommand{\imgh}{1277.0}
\centering
\foreach \method/\stamp in {GPMVS~\cite{hou2019multi}/gpmvs01.jpeg, Atlas~\cite{murez2020atlas}/atlas00.jpeg, Vortx~\cite{stier2021vortx}/vortx12.jpeg, NeuralRecon~\cite{sun2021neuralrecon}/neucon05.jpeg, Ours/ours07.jpeg, {Ours (High Reso)}/ours_high08.jpeg, GT/gt04.jpeg} {
	\begin{subfigure}[t]{\w}
		\caption*{\textbf{\method}}
		\begin{tikzpicture}
			\node[anchor=south west,inner sep=0] (image) at (0,0) {\includegraphics[width=\columnwidth]{figures/shownmesh/scene0806_00/\stamp}};
			\begin{scope}[x={(image.south east)},y={(image.north west)}]
				\draw[red] (0.33, 0.45) rectangle (0.59, 0.75);
			\end{scope}
		\end{tikzpicture}
	    \par\smallskip
		\adjincludegraphics[cfbox=red 2pt, width=0.9\columnwidth, trim={{0.33\width} {0.45\height} {0.41\width} {0.25\height}},clip]{figures/shownmesh/scene0806_00/\stamp}
	\end{subfigure} \hspace{-9pt} 
}
 

\par\smallskip
\par\noindent\hdashrule{\textwidth}{0.4pt}{0.8pt}
\par\smallskip
\centering
\foreach \seq/\stamp in {gpmvs02.jpeg, atlas01.jpeg, vortx00.jpeg, neucon04.jpeg, ours05.jpeg, ours-high06.jpeg, gt03.jpeg} {
	\begin{subfigure}[t]{\w}
		\begin{tikzpicture}
			\node[anchor=south west,inner sep=0] (image) at (0,0) 	{\includegraphics[width=\columnwidth]{figures/shownmesh/scene0750_00/\stamp}};
			\begin{scope}[x={(image.south east)},y={(image.north west)}]
				\draw[red]   (0.13, 0.29) rectangle (0.35, 0.52);
			\end{scope}
		\end{tikzpicture}
		\par\smallskip
		\adjincludegraphics[cfbox=red 2pt, width=\columnwidth, trim={{0.13\width} {0.29\height} {0.65\width} {0.48\height}},clip]{figures/shownmesh/scene0750_00/\stamp}
	\end{subfigure} \hspace{-9pt}
}

\par\smallskip
\par\noindent\hdashrule{\textwidth}{0.4pt}{0.8pt}
\par\smallskip
\centering
\foreach \seq/\stamp in {gpmvs02.jpeg, atlas01.jpeg, vortx06.jpeg, neucon03.jpeg, ours04.jpeg, ours-high05.jpeg, gt00.jpeg} {
	\begin{subfigure}[t]{\w}
		\begin{tikzpicture}
			\node[anchor=south west,inner sep=0] (image) at (0,0) 	{\includegraphics[width=\columnwidth]{figures/shownmesh/scene0747_00/vis/\stamp}};
			\begin{scope}[x={(image.south east)},y={(image.north west)}]
				\draw[red]   (0.16, 0.33) rectangle (0.35, 0.6);
			\end{scope}
		\end{tikzpicture}
			\par\smallskip
		\adjincludegraphics[cfbox=red 2pt, width=\columnwidth, trim={{0.16\width} {0.33\height} {0.65\width} {0.4\height}},clip]{figures/shownmesh/scene0747_00/vis/\stamp}
	\end{subfigure} \hspace{-9pt}
}

\caption{
	 qualitative results on Scannet test sequences~\cite{dai2017scannet}. We also zoom in the region in red rectangles for clear views. The proposed method, \textit{Ours}, can recover more 3D structures than the incremental feature-volume-based method NeuralRecon~\cite{sun2021neuralrecon}. With a higher resolution, \textit{Ours (High Reso)} is able to recover more fine-detailed structures. 
}
\vspace{-1em}
\label{fig:meshes} 
\end{figure*}

\subsection{Datasets and Metrics}
For all the evaluations, we use the ScanNet dataset~\cite{dai2017scannet} for training, which consists of 1513 RGBD sequences collected in 707 indoor scenes. We follow the official training and test splits, which are 1201 and 100 sequences, respectively. Besides the ScanNet test split, we also test the ScanNet-trained network zero-shot on TUM-RGBD~\cite{sturm2012benchmark} (13 sequences following~\cite{stier2021vortx}), and our own collected dataset without any finetuning.
For 3D metrics, we evaluate the final reconstructed surface mesh extracted from the predicted TSDF volume against the officially provided mesh on ScanNet, and we generate our own meshes on TUM-RGBD and our own collected datasets through conventional TSDF fusion using the ground truth depth. \rev{Following the evaluation protocol~\cite{murez2020atlas, sun2021neuralrecon} exactly, we calculate 3D metrics, including accuracy, completeness, precision, recall, and F-score, across uniformly sampled points with a 2-centimeter resolution from dense meshes. For computing these 3D metrics, a distance threshold of 5 centimeters was used.} We regard the F-score to be the most representative metric to reflect the quality of 3D reconstruction, since it is involved with both precision and recall.
\rev{}
Regarding 2D metrics, we evaluate the rendered depth maps at all the image views for the feature volume-based scene reconstruction methods, against the provided raw depth maps with a truncation of 10 meters. The MVS methods predicting dense depth maps directly allow for handy evaluations.

\subsection{Training Details}

We use ScanNet dataset~\cite{dai2017scannet} for training. 
%
Ground truth TSDF volumes are generated from raw depth maps and given camera poses by conventional TSDF fusion~\cite{curless1996volumetric, newcombe2011kinectfusion}. Note that we discard all the depths over $3m$ like existing methods~\cite{sun2021neuralrecon, stier2021vortx}. 
Before training the feature volume pipeline, we need to fine-tune the lightweight GPMVS~\cite{hou2019multi} depth prediction network for 12 epochs, then train the depth variance prediction network for 4 epochs from randomly initialized weights. Note that, GPMVS is frozen in subsequent training, while the weights of the variance network are kept updated. 
We have two phases for training the feature volume pipeline for incremental reconstruction from monocular videos.
At the first phase, we train the fragment-wise reconstruction network, regressing TSDF from feature volume for 20 epochs.
The network learns how to predict TSDF from aggregated features in a fragment volume with size $3.84m \times 3.84m \times 3.84 m$. Finally, we train the network at the second phase with the GRU fusion network together for 30 epochs. 
Adam optimizer with $\beta_1 = 0.9, \beta_2 = 0.999$ is adopted for training the networks, and the learning rate is $1e-3$ at the beginning and decreased by half at epochs $12, 24, 48$ in the two-phase training.
The finest voxel resolution is $4cm$ by default, and is reduced to $2cm$ for our high-resolution variant. The batch size is $32$ by default, and $8$ for the higher resolution variant.
%

\subsection{Evaluation on ScanNet}
The evaluation results with both 3D metrics and 2D metrics are shown in Table~\ref{tab:scannet-3d} and Table~\ref{tab:scannet-2d}, respectively. We compare the proposed method, \textit{Ours}, with state-of-the-art traditional structure-from-motion method, COLMAP~\cite{schonberger2016pixelwise}, multi-view stereo networks including MVDepthNet~\cite{wang2018mvdepthnet}, DPSNet \cite{im2019dpsnet}, and GPMVS \cite{hou2019multi}, as well as all the open-source feature-volume-based methods including Atlas \cite{murez2020atlas},  Vortx \cite{stier2021vortx}, and NeuralRecon \cite{sun2021neuralrecon}.
Our method with high-resolution feature volume is named \textit{Ours (High Reso)}. It should be noted all the compared methods are evaluated with the same protocol.
%
%
All compared deep-learning-based methods have been trained or fine-tuned on the ScanNet dataset. The results of  COLMAP, MVDepthNet, and DPSNet are taken from~\cite{murez2020atlas}, while GPMVS, Atlas, Vortx, and NeuralRecon are evaluated by ourselves.

The proposed incremental feature volume-based method outperforms the existing incremental method, NeuralRecon, regarding the representative 3D reconstruction metric -- the F-score. The advantages on F-score are mainly from the higher recall, while the proposed method and NeuralRecon have very similar precision. With the incorporation of MVS depth and depth uncertainty in our method, more structures can be recovered compared to the pure feature volume-based NeuralRecon.
When the voxel size of feature volume is decreased from $4cm$ to $2cm$, dubbed as \textit{Ours (High Reso)}, the overall quality of reconstruction can be further improved. 
We attempted to train NeuralRecon~\cite{sun2021neuralrecon} with a high-resolution feature volume as well for a direct comparison, but the training network was unable to fit in GPU memory (see Sec. \ref{subsec:memruntime}).
It should also be noted that both Atlas~\cite{murez2020atlas} and  Vortx~\cite{stier2021vortx} are offline methods, and have the access to full batch data with global context before predicting TSDF values from features, and they are expected to exhibit better performance than our real-time incremental method. Actually, \textit{Ours} as the real-time incremental method has very close performance to Atlas~\cite{murez2020atlas} and slightly better performance than traditional offline method COLMAP~\cite{schonberger2016pixelwise}. 

We also show qualitative results of the meshes generated from different methods in Fig.~\ref{fig:meshes}.
%
The MVS method, GPMVS~\cite{hou2019multi}, which is also leveraged in our pipeline to guide the feature allocation, suffers from significant artifacts. With the feature volume-based regularization and denoising, our method is able to remove the main artifacts in MVS reconstruction. This also verifies that the predicted depth and depth uncertainty for feature allocation only near the surface voxels are reasonable, and have provided the rough shape sufficient for the proceeding feature volume-based reconstruction.
It is found that \textit{Ours} and \textit{Ours (High Reso)} have the potential to recover more details about objects, such as the table leg, toilet, and chair. 
Atlas~\cite{murez2020atlas} and  Vortx~\cite{stier2021vortx} are prone to recover more complete walls and floors, but their predictions can be over-smoothed to miss details of objects, and over-filled with hallucinated structures.

\begin{figure} [!thb]
\vspace{-1em}
\color{blue}
	\newcommand{\w}{.165\textwidth}
	\centering 
\begin{subfigure}[t]{\w}
		\caption*{\textbf{NeuralRecon~\cite{sun2021neuralrecon}}}
		\includegraphics[width=\columnwidth]{
  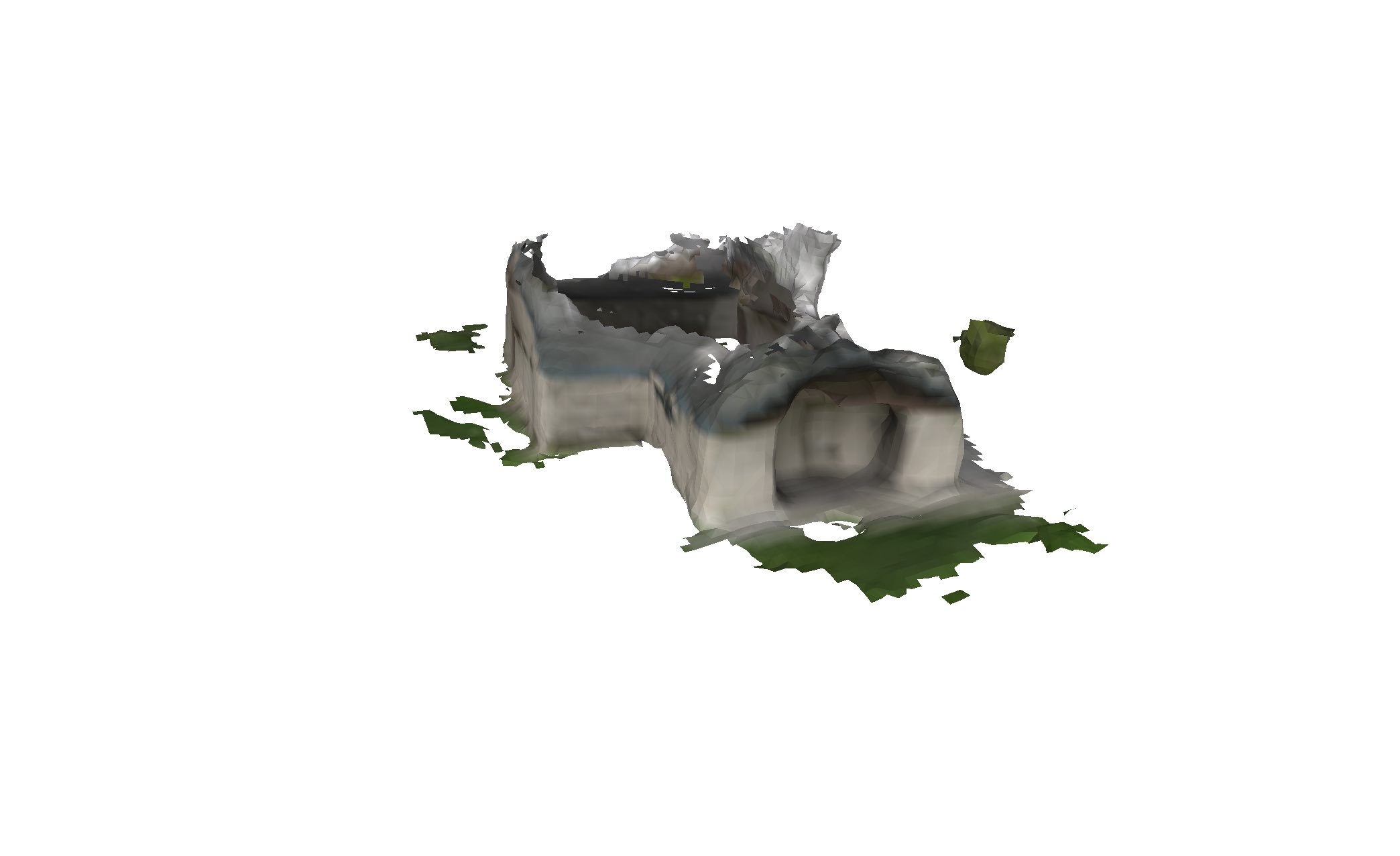
  }
\end{subfigure}
 \hspace{-14pt}
\begin{subfigure}[t]{\w}
		\caption*{\textbf{Ours}}
		\includegraphics[width=\columnwidth]{
  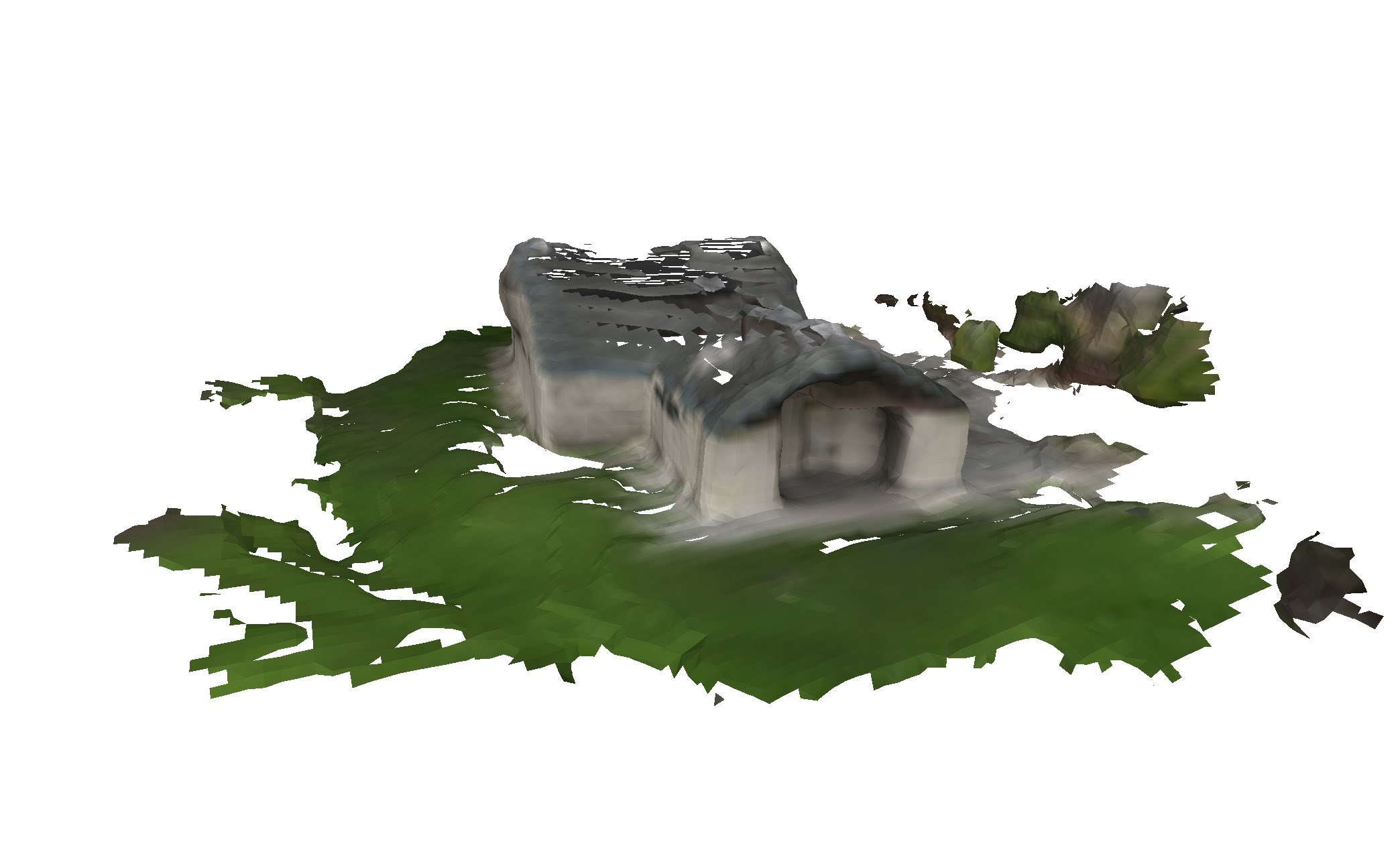
   }
\end{subfigure}
 \hspace{-14pt}
\begin{subfigure}[t]{\w}
		\caption*{\textbf{Ours (High Reso)}}
		\includegraphics[width=\columnwidth]{
  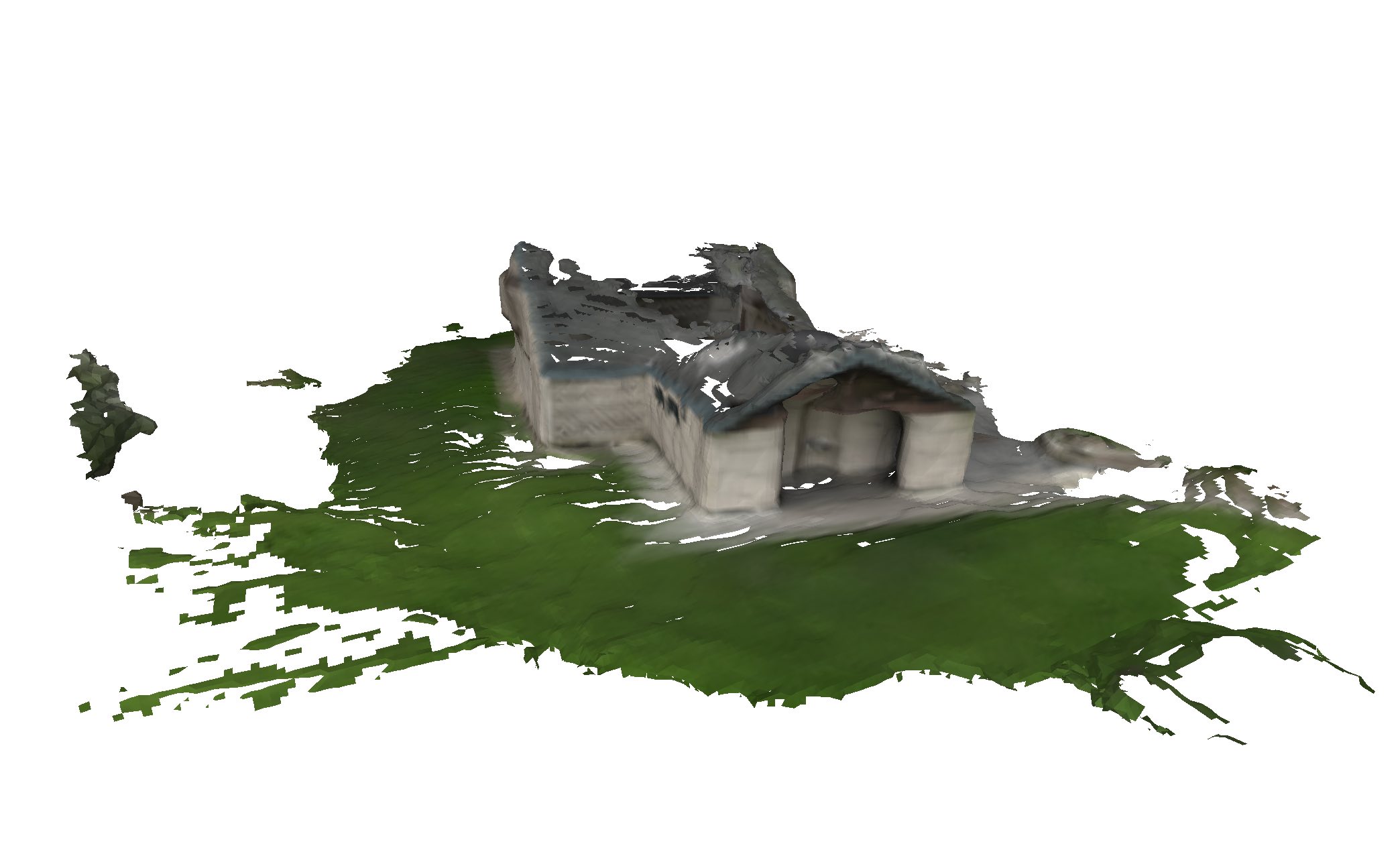
  }
\end{subfigure} \\
\vspace{-1em}
\begin{subfigure}[t]{\w}
		\includegraphics[width=\columnwidth]{
  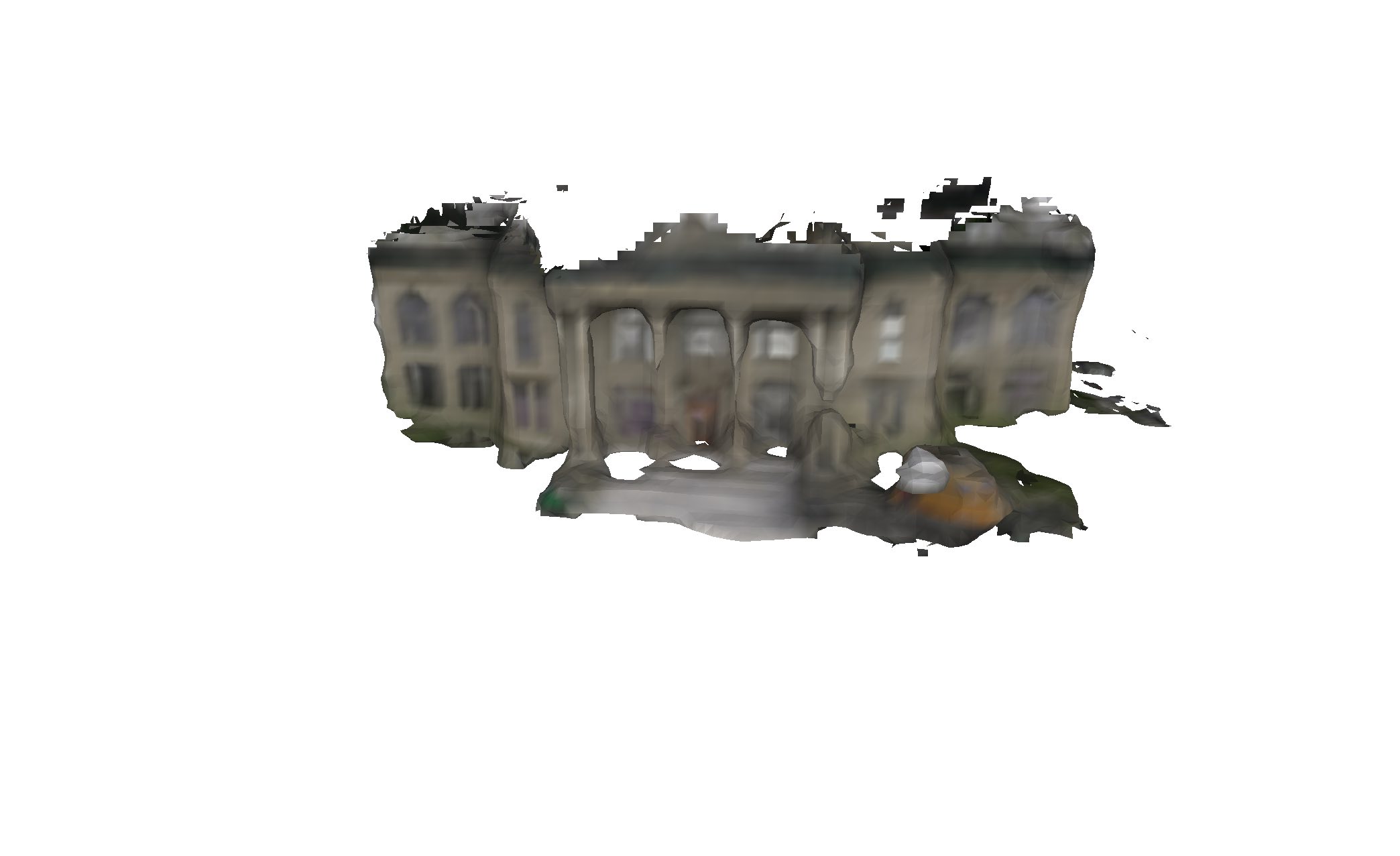
  }
\end{subfigure}
 \hspace{-24pt}
\begin{subfigure}[t]{\w}
		\includegraphics[width=\columnwidth]{
  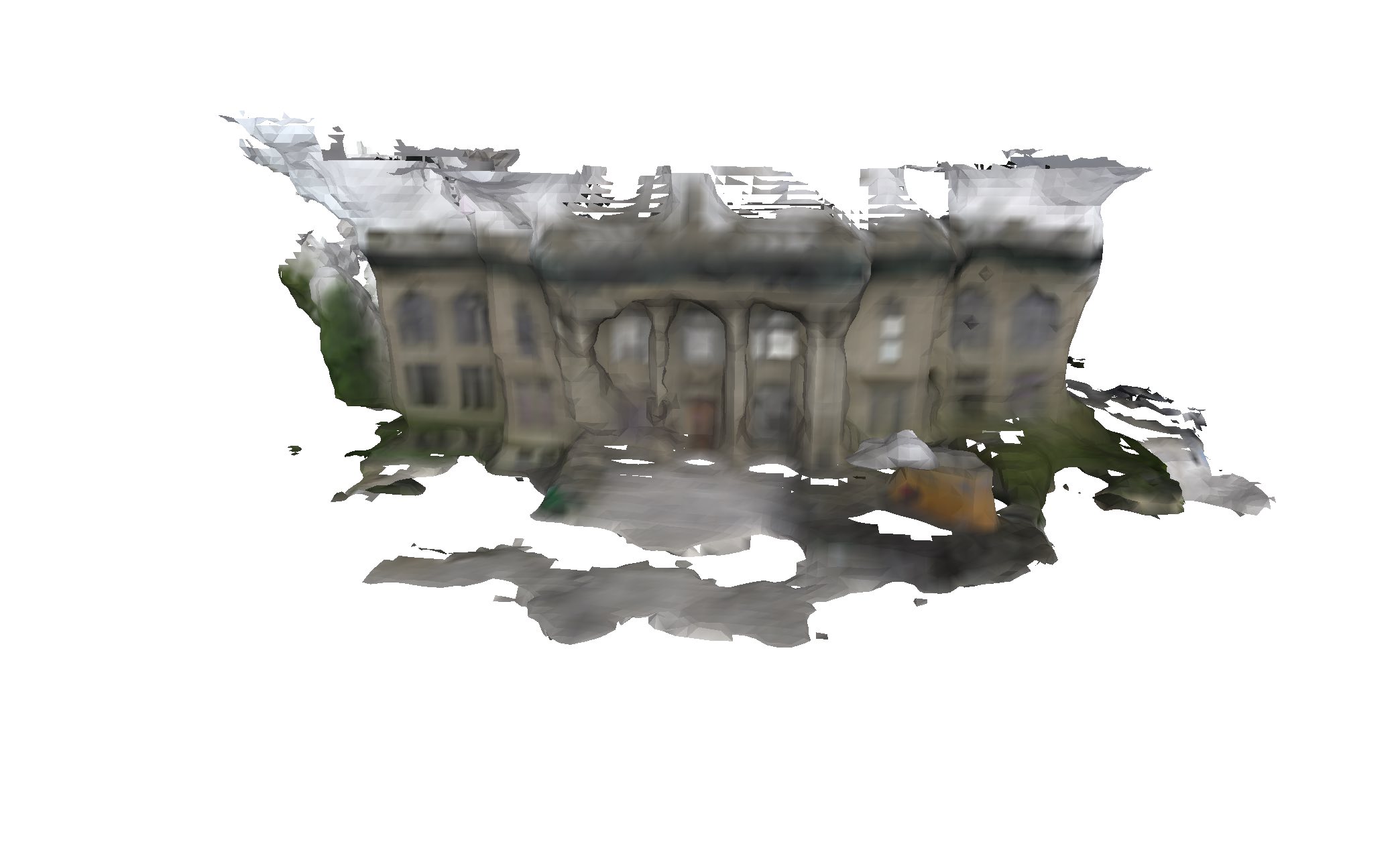
   }
\end{subfigure}
 \hspace{-24pt}
\begin{subfigure}[t]{\w}
		\includegraphics[width=\columnwidth]{
  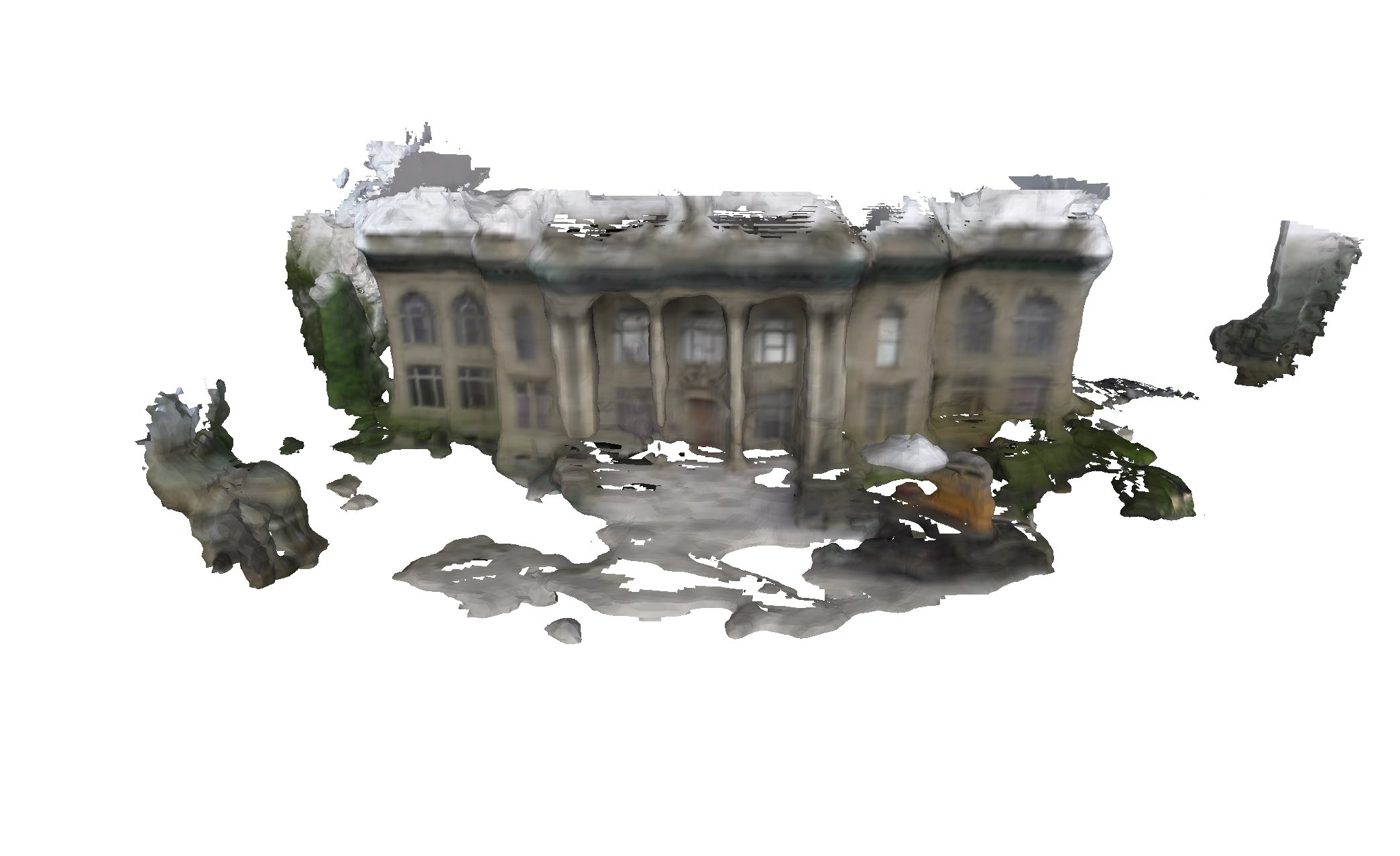
  }
\end{subfigure}
\vspace{-1em}
\caption{Qualitative results in outdoor large-scale scenarios on Tanks \& Temples~\cite{tanktemple2017} sequences. 
} 
    \label{fig:meshes_tanktemple}
    \vspace{-1em}
\end{figure}


\subsection{Generalization on TUM-RGBD}
To examine the generalization of the proposed method, we also conduct comparisons on TUM-RGBD dataset~\cite{sturm2012benchmark}. We only compare to the incremental feature volume-based methods. We follow exactly the identical keyframing strategy as the one on Scannnet dataset. The 3D evaluation metrics are shown in Table~\ref{tab:tumrgbd}. We can find that \textit{Ours} outperforms NeuralRecon~\cite{sun2021neuralrecon}. Our high-resolution variant has obvious advantages over the default resolution. 

%

\begin{table}[tbh]
	\centering 
	\caption{Evaluated on TUM-RGBD dataset.}
	\label{tab:tumrgbd}
	\resizebox{\columnwidth}{!}{
		\begin{tabular}{cccccc}
			\Xhline{3\arrayrulewidth}
			Method       & Comp $\downarrow$ & Acc $\downarrow$   & Recall $\uparrow$ & Prec $\uparrow$   & F-score  $\uparrow$  \\ \hline
			\rowcolor[gray]{0.902}
			NeuralRecon \cite{sun2021neuralrecon} & 0.285   & 0.101 & 0.109& 0.391& 0.169  \\
			\rowcolor[gray]{0.902}
			Ours & 0.252 & 0.123 & 0.133 & 0.369 & 0.195  \\
			\rowcolor[gray]{0.902}
			\rev{Ours (High Reso)}& \textbf{0.192}& \textbf{0.084}& \textbf{0.199}& \textbf{0.586}& \textbf{0.295} \\
			\Xhline{3\arrayrulewidth}
		\end{tabular}
	}
	\vspace{-2em}
\end{table}

\subsection{Generalization on Tanks \& Temples }
\rev{We further conduct evaluations in two large-scale outdoor scenarios, ``Barn" and ``Courthouse", using the Tanks \& Temples dataset~\cite{tanktemple2017} without any fine-tuning of the network weights trained on the indoor ScanNet dataset~\cite{dai2017scannet}.
	Our proposed method is designed to maintain sparsity in the feature volume, making it well-suited for large-scale volumetric reconstruction where most voxel grids are physically empty. In contrast, allocating features to a large number of voxels in dense volume can become computationally intractable. 
	In order to run NeuralRecon~\cite{sun2021neuralrecon} successfully on these large-scale scenarios, the poses of images for all the compared methods are scaled down to be 5 to 10 times smaller.  
	%
	The dense reconstruction results are qualitatively depicted in Fig.~\ref{fig:meshes_tanktemple}. 
	Our method demonstrates strong generalization capabilities in outdoor large-scale scenarios, visibly outperforming NeuralRecon with much more desirable reconstruction results.  Moreover, our high-resolution reconstruction can recover finer details and more accurate scene structures.
}

\subsection{Generalization on Self-Collected Data}
\begin{figure*} [thb]
	\centering
	\newcommand{\w}{.20\textwidth}
	\centering 
	\begin{subfigure}[t]{\w}
		\caption*{\textbf{NeuralRecon~\cite{sun2021neuralrecon}}}
		\includegraphics[width=0.85\columnwidth]{
			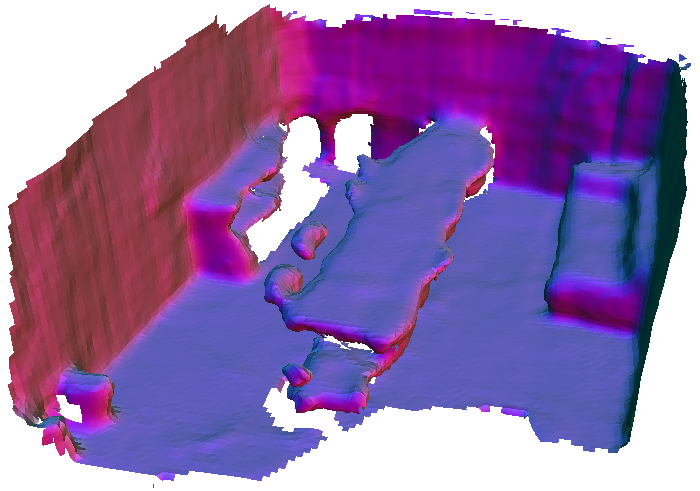
		}
	\end{subfigure}
	\hspace{-7pt}
	\begin{subfigure}[t]{\w}
		\caption*{\textbf{Ours}}
		\includegraphics[width=0.85\columnwidth]{
			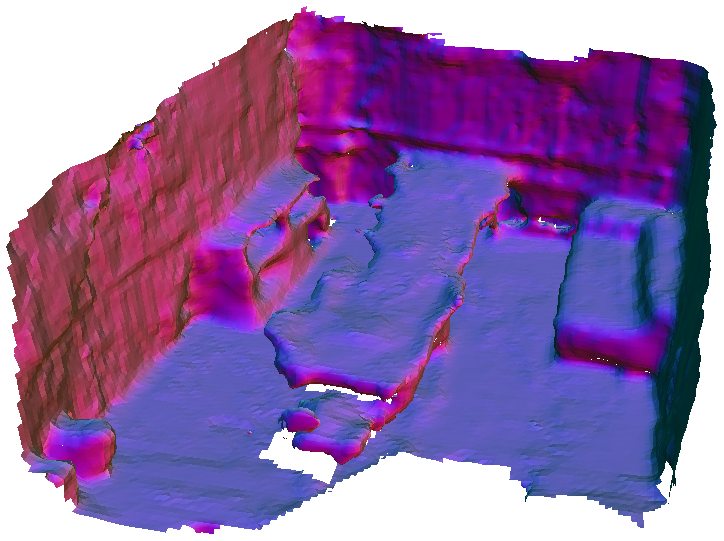
		}
	\end{subfigure}
	\hspace{-7pt}
	\begin{subfigure}[t]{\w}
		\caption*{\textbf{Ours (High Reso)}}
		\includegraphics[width=0.85\columnwidth]{
			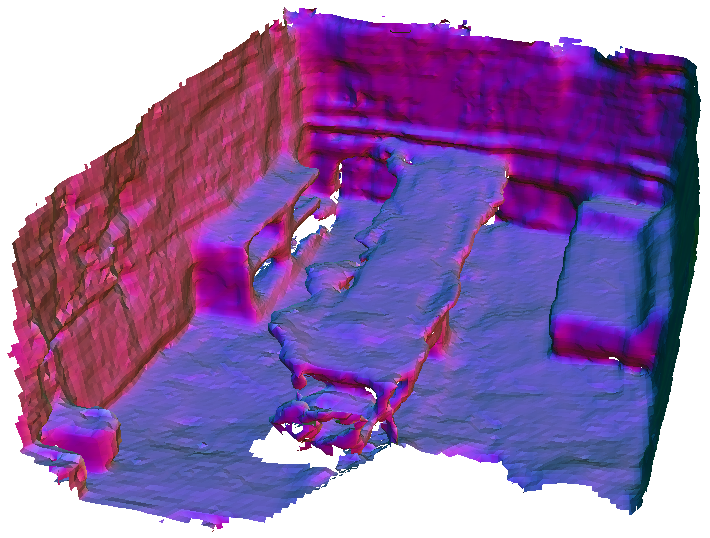
		}
	\end{subfigure}
	\hspace{-7pt}
	\begin{subfigure}[t]{\w}
		\caption*{\textbf{GT}}
		\includegraphics[width=0.85\columnwidth]{
			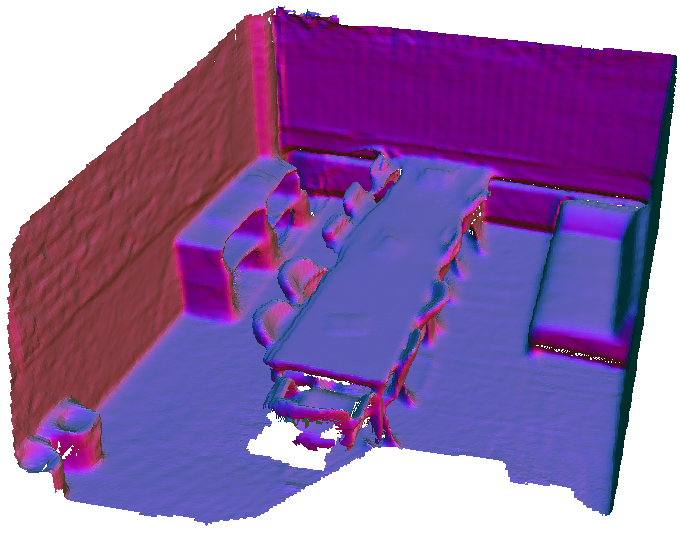
		}
	\end{subfigure}
	\hspace{-7pt}
	\begin{subfigure}[t]{\w}
		\caption*{\textbf{RGB}}	
		\includegraphics[width=0.85\columnwidth]{%
			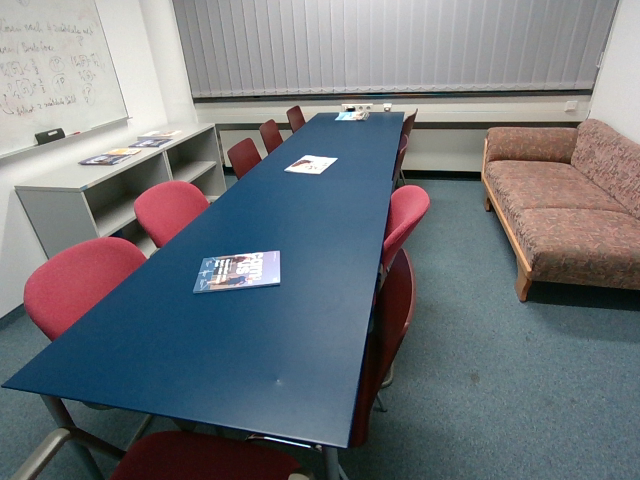}
	\end{subfigure}
	\caption{Qualitative results of a representative sequence from our self-collected dataset.} 
	\label{fig:meshes_realsense}
\end{figure*}

We also collect our own dataset with 8 sequences in indoor scenarios by the Realsense D455 RGBD camera~\footnote{https://www.intelrealsense.com/depth-camera-d455}. We record the grayscale stereo images, RGB images, depth maps, and IMU data streamed from the camera.
The stereo images and IMU data are fed into a visual-inertial SLAM system, OKVIS 2.0~\cite{leutenegger2022okvis2}, for getting accurate 6DoF poses. 
A snapshot of a typical scenario, and the reconstructed meshes from compared methods are shown in Fig.~\ref{fig:meshes_realsense}, where we can easily find that our methods can recover more geometric details than NeuralRecon~\cite{sun2021neuralrecon}. The quantitative 3D metric evaluations are also shown in Table~\ref{tab:realsense}.

%

\begin{table}[tbh]
	\centering 
	\caption{Evaluated on our own collected data.}
	\label{tab:realsense}
	\resizebox{\columnwidth}{!}{
		\begin{tabular}{cccccc}
			\Xhline{3\arrayrulewidth}
			Method       & Comp $\downarrow$ & Acc $\downarrow$   & Recall $\uparrow$ & Prec $\uparrow$   & F-score  $\uparrow$   \\ \hline
			\rowcolor[gray]{0.902}
			NeuralRecon \cite{sun2021neuralrecon} & 0.205& \textbf{0.034}& 0.215& \textbf{0.774}& 0.335 \\
			\rowcolor[gray]{0.902}
			Ours & 0.144& 0.040& 0.255& 0.711& 0.374  \\
			\rowcolor[gray]{0.902}
			\rev{Ours (High Reso)} & \textbf{0.143}& 0.045& \textbf{0.268}& 0.683& \textbf{0.385} \\
			\Xhline{3\arrayrulewidth}
		\end{tabular}
	}
	\vspace{-2em}
\end{table}

\subsection{Memory and Runtime} \label{subsec:memruntime}
%
%
We conducted runtime and memory evaluations of the inference stage on a desktop computer equipped with an RTX5000@16GB GPU and 8 Intel i7-11700k CPU Cores@3.6GHz. 
In Table~\ref{tab:mem_runtime}, we report the averaged time taken for MVS depth recovery, feature encoding, feature aggregation, TSDF-Fusion of MVS depths, 3D sparse CNN, GRU fusion, and the total processing time for a volume chunk with 9 keyframes. The experiment is conducted on a typical large-scale indoor sequence of ScanNet at the inference stage. 
%
Despite incorporating MVS depth and self-attention,
our method can run incremental reconstruction in real-time at 12.70 keyframes per second \rev{, with a slightly increased memory footprint compared to NeuralRecon.}
\rev{At the high resolution, our method runs at 5 keyframes per second.}
Although this is admittedly slower than NeuralRecon, which can run at 41 keyframes per second on the same device, it should be noted that our method is still real-time capable
since keyframes in a typical real-time SLAM system (e.g., \cite{leutenegger2022okvis2})
are created at a far lower frequency than the framerate. 

Our method is additionally highly memory-efficient in terms of voxel allocation.
At the inference stage, compared with the `dense' feature volume of NeuralRecon, the numbers of the non-empty voxel with allocated features in our sparse volume are significantly decreased by $67.71\%, 48.24\%, 30.71\%$  at three levels respectively.
The feature volume in NeuralRecon is initially dense, and is sparsified by the network from coarse to fine levels.
This means that, in the initial phases of training, before the network learns to sparsify well, a nearly dense volume makes its way through the sparse 3D convolution network.
In our experiments, we did not train NeuralRecon at the high resolution due to this issue since the required GPU memory exceeded 42GB -- even with batch size 1.
In contrast, due to our MVS-guided sparse volume allocation, which makes feature volumes sparse at all three levels, we are able to train at the higher resolution even with the memory-intensive attention mechanism involved, consuming GPU memory around 28GB during training.

\begin{table*}[tbh]
	\centering 
	\caption{Mean of runtime (Second), memory consumption (GB), and number of nonempty voxels in three-level feature volumes, during the reference on a typical sequence of ScanNet.}
	\label{tab:mem_runtime}
	\resizebox{0.95\textwidth}{!}{
		\begin{tabular}{ccccccccC{0.3cm}ccccc}\cmidrule[3\arrayrulewidth]{1-8}\cmidrule[3\arrayrulewidth]{10-14}
			Method       & MVS Dep. & Feat. Enc. & TSDF-Fusion  & Feat. Agg. & 3D SpCNN   & GRU Fusion & Total & &  Kf./Sec.& Voxels L0 & Voxels L1& Voxels L2 & Memory (GB)\\ \cline{1-8}\cline{10-14}
			NeuralRecon \cite{sun2021neuralrecon} & - & 0.037 & - & 0.012 & 0.096 & 0.063 & 0.219& & 41.10 & 4307.436 & 11893.692 & 46665.385 & 0.988 \\
			Ours & 0.377& 0.025& 0.061& 0.099& 0.060& 0.057& 0.711 & &12.66 & 1390.744 & 6156.615  & 32336.641 & 1.68 \\
			Ours (High Reso) & 0.384& 0.025 & 0.181& 0.573& 0.225& 0.315 & 1.779& &5.06 & 6962.026 &39470.615 &  230968.897 & 9.06 \\
			\cmidrule[3\arrayrulewidth]{1-8}\cmidrule[3\arrayrulewidth]{10-14}
		\end{tabular}
	}
	\vspace{-1em}
\end{table*}


\subsection{Ablation Study}
To examine the effectiveness of our design choices, we conduct ablation studies on the ScanNet dataset at the default resolution. The results are reported in Table~\ref{tab:ablation-3d}. We first examine our method without sparsification. 
%
%
In this case, the F-score is a bit higher, while running our method without the sparsification incurs a higher memory consumption due to more voxels needing to be allocated.
We further ablate our method by removing the predicted depth uncertainty for feature allocation~-- allocating features within a constant distance of 5 voxels around the surface recovered from the predicted MVS depth. Noticeably, our full method has better performance due to the probabilistic feature allocation accounting for the uncertainty of predicted MVS depth.
It is also clear from the table that our choice of feature augmentation using the TSDF values and weights generated from trivial TSDF fusion of MVS depth, as well as our choice of self-attention for feature aggregation, have significant effects on the system performance.
The hints from MVS TSDF values and weights can guide the network for better convergence. The attention mechanism for feature aggregation from multiple views enables selectively absorbing informative deep features for 3D structure recovery.
For the last ablation study in Table~\ref{tab:ablation-3d}, we investigate whether it is possible to reduce the number of parameters by sharing the feature encoder of the MVS Network with the 2D CNN for feature extraction (see Fig.~\ref{fig:framework}), instead of using separate feature encoders for the two modules in our design choice. Notably, the performance is significantly deteriorated due to the fact that MVS and feature volume fusion \rev{require different features}.

\begin{table}[!t]
	\centering 
	\caption{Ablation study: 3D geometry metrics evaluated on ScanNet test split.}
	\label{tab:ablation-3d}
	\resizebox{\columnwidth}{!}{
		\begin{tabular}{cccccc}
			\Xhline{3\arrayrulewidth}
			Method       & Comp $\downarrow$ & Acc $\downarrow$   & Recall $\uparrow$ & Prec $\uparrow$   & F-score  $\uparrow$    \\ \hline
			Ours (High Reso) & 0.116 & 0.056 & 0.525 & 0.675 & 0.589 \\
			Ours & 0.110 & 0.058 & 0.505 & 0.665 & 0.572  \\ \hline
			Ours: w/o sparsification & 0.123& 0.047& 0.487&0.713& 0.577 \\
			Ours: w/o depth uncer. &  0.111& 0.061& 0.496& 0.651& 0.561 \\
			Ours: w/o tsdf augment. & 0.120& 0.063& 0.472& 0.637& 0.540 \\
			Ours: w/o attention & 0.119& 0.073& 0.451& 0.592& 0.511\\
			Ours: w/o sep. feat. enc.& 0.121& 0.065& 0.463& 0.625& 0.530 \\
			\Xhline{3\arrayrulewidth}
		\end{tabular}
	}
	\vspace{-2em}
\end{table}

\subsection{Limitations and Discussions}
Our proposed method leverages \rev{MVS} depth to guide incremental feature volume-based reconstruction. MVS can provide rough locations of the physical surface and enable sparse allocation, while feature volume-based fusion can further regularize, refine, and denoise the recovered 3D structures from MVS depth. With the MVS guidance, more geometric details can be recovered. However, it pays the cost that local smoothness can be slightly degraded. Since the sparse feature allocation is critical for our proposed method, if the MVS network fails to predict a depth distribution surrounding the true physical surface, the feature volume-based pipeline can not recover appreciable 3D structures from empty voxels.

\section{Conclusion and Future Work}\label{sec:conc}
We presented a real-time incremental 3D dense reconstruction method from monocular videos based on MVS, attention mechanism, sparse 3D CNN, and GRU fusion. Predicted depth maps and uncertainties from MVS neural networks provide an initial guess of the physical surface locations in feature volume. Then feature volume-based pipeline temporally fuses the deep features into the sparsified feature volume to refine 3D geometries and impose local smoothness and 3D priors learned from data. 
The proposed method is demonstrated to perform accurate 3D dense reconstruction on several datasets, and can scale up to high-resolution reconstruction due to its memory-efficient nature in terms of sparse feature allocation. 

\section*{Acknowledgment}
We thank Noah Stier~\cite{stier2021vortx}, Jiaming Sun, and Yiming Xie~\cite{sun2021neuralrecon} for discussions about the evaluations of baselines.

{
\def\bibfont{\scriptsize}
\printbibliography
%
}

\end{document}